\documentclass[runningheads]{llncs}
\usepackage{graphicx}
\usepackage{times}
\usepackage{epsfig}
\usepackage{subfig}
\usepackage{amssymb}
\usepackage{cite}
\usepackage{framed,multirow}
\usepackage[utf8]{inputenc}
\usepackage[english]{babel}

\usepackage[none]{hyphenat}
\emergencystretch=\maxdimen
\newcommand{\etal}{\textit{et al}.}

\begin{document}
\title{Exploring Modulated Detection Transformer as a Tool for Action Recognition in Videos}
\titlerunning{Exploring MDETR as a Tool for Action Recognition}

\author{Tomás Crisol\inst{1}\thanks{These authors contributed equally}
\and Joel Ermantraut\inst{1}\thanks{These authors contributed equally}
\and Adrián Rostagno\inst{1}
\and Santiago L. Aggio\inst{1, 2}
\and Javier Iparraguirre\inst{1}}

\authorrunning{T. Crisol et al.}

\institute{
Universidad Tecnológica Nacional, Facultad Regional Bahía Blanca, Argentina
\email{tomascrisol12,joelermantraut@gmail.com} \\
\email{arostag@frbb.utn.edu.ar} \\
\email{j.iparraguirre@computer.org} 
\and CONICET, Bahía Blanca, Argentina\\
\email{slaggio@criba.edu.ar} 
}

\maketitle       
\begin{abstract}
During recent years transformers architectures have been growing in popularity. Modulated Detection Transformer (MDETR) is an end-to-end multi-modal understanding model that performs tasks such as phase grounding, referring expression comprehension, referring expression segmentation, and visual question answering. One remarkable aspect of the model is the capacity to infer over classes that it was not previously trained for. In this work we explore the use of MDETR in a new task, action detection, without any previous training. We obtain quantitative results using the Atomic Visual Actions dataset. Although the model does not report the best performance in the task, we believe that it is an interesting finding. We show that it is possible to use a multi-modal model to tackle a task that it was not designed for. Finally, we believe that this line of research may lead into the generalization of MDETR in additional downstream tasks.

\end{abstract}

\section{Introduction}
Transformers architectures have been increasing in popularity among the machine learning community \cite{khan2021transformers}. Initially, this type of architecture emerged in the natural language processing space \cite{vaswani2017attention}. However, it is possible to observe a rapid expansion in other modalities such as computer vision \cite{khan2021transformers}. Recently, multi-modal transformers enabled the possibility to process images and text using a single model. Additionally, video understanding tasks were tackled by transformers models such as the work proposed by Wang \etal \cite{wang2021long}.

Modulated Detection Transformer (MDETR) \cite{kamath2021mdetr} is a multi-modal transformer. The architecture accepts an image and text as input, and it can be trained on multiple downstream tasks. One particular task is visual question answering. Although the initial design of the model does not target video understanding, we used MDETR as an action recognition model. Without any additional training, we evaluated the performance of the model on an action recognition dataset. Naturally, the results are not the best reported. However, we found it valuable to assess the use of a multi-modal transformer in tasks that it was not designed for. It is important to highlight that no previous training was performed before the evaluation.

The Atomic Visual Actions (AVA)\cite{gu2018ava} dataset consists of a collection of 430 videos annotated with 80 visual actions. It contains 1.58M action labels associated with a bounding box. Since MDETR provides coordinates that are related to the output, it is possible to ask a question and get an answer with the related area of interest. We ran experiments on AVA and we obtained quantitative results. Additionally, all reported findings were published in an open repository \footnote{\url{https://github.com/BHI-Research/AVA\_MDETR}}. Next section explores the related work. In section \ref{section-results} quantitative results are presented. Finally, conclusions are stated in section \ref{section-conlusions}.

\section{Related Work}
\subsection{MDETR}
Modulated Detection Transformer (MDETR) \cite{kamath2021mdetr}, performs object detection in conjunction with language understanding. The concept enables end-to-end multi-modal understanding. The model relies only on text and the aligned boxes in an image. Unlike previous detection methods, MDETR detects concepts from free text and generalizes to unseen combinations of categories and attributes.
Quantitative results reported by MDETR authors are outstanding in four tasks. Reported categories were phase grounding, referring expression comprehension, referring expression segmentation, and visual question answering. Given a collection of videos, we sampled the clips and extracted 1 frame per second. Using visual question answering, we asked for actions and measured the output of the system.

As any transformer, MDETR was trained in two stages, the pre-training and the downstream tasks. During pre-training, the model ingested a combination of Flickr30k \cite{plummer2015flickr30k}, MS COCO \cite{lin2014microsoft} and Visual Genome (VG) \cite{krishna2017visual} datasets. In the case of the visual question answering task, GQA \cite{hudson2019gqa} was the selected dataset. During inference, the model reads a tensor of linear image features and a tensor of linear text features. Then, the input is concatenated and fed into an encoder. Depending on the task, the decoder presents some variation. In the case of question answering, object queries and specific queries are fed into the decoder. As a result, the decoder provides new object positions and answers to the queries. 


\subsection{AVA}

AVA \cite{gu2018ava} dataset is a person centered corpus, annotated at a 1 Hz sample rate. Every person is located using a bounding box and the labels correspond  to actions related to the pose, interactions with objects, and interactions with other persons. The temporal context of the annotation is centered $\pm$ 1.5 second around the keyframe. This ``brief'' time lapse gives the name to the dataset. Annotations in the dataset are precise and the number of labels reaches 1.58 million. 

Multiple metrics are available in AVA. Intersection-over-union (IoU) is reported at frame level and at video level. In the case of frame level, the metric is built following the standard protocol used by the PASCAL VOC challenge \cite{everingham2015pascal}. Average precision (AP) is computed using an IoU threshold of 0.5. Mean Average Precision (mAP) is the average of AP over all classes. In this work, AP is reported.

\section{Results}
\label{section-results}

Experiments reported on this work were obtained using the original MDETR model and the AVA actions dataset v2.2. Since the model takes images and text as input, we sampled the videos at 1 Hz to obtain the frames. Regarding actions, we created a collection of questions that ask for the actions vocabulary annotated in AVA. For each frame extracted from the dataset, we asked all the questions available. The output of the model was saved in a CSV file. Afterwards, we obtained quantitative results. 

\begin{figure}%
 \centering
 \subfloat[]{{\includegraphics[width=.46\textwidth]{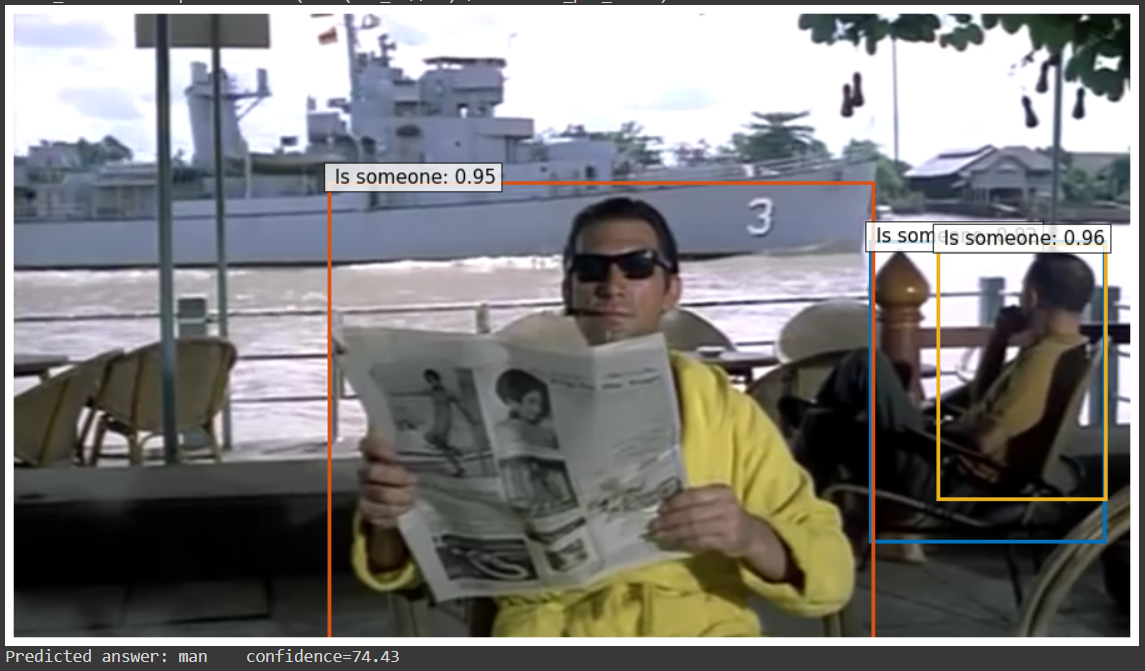} \label{fig-all-cus}}}%
 \qquad
 \subfloat[]{{\includegraphics[width=.46\textwidth]{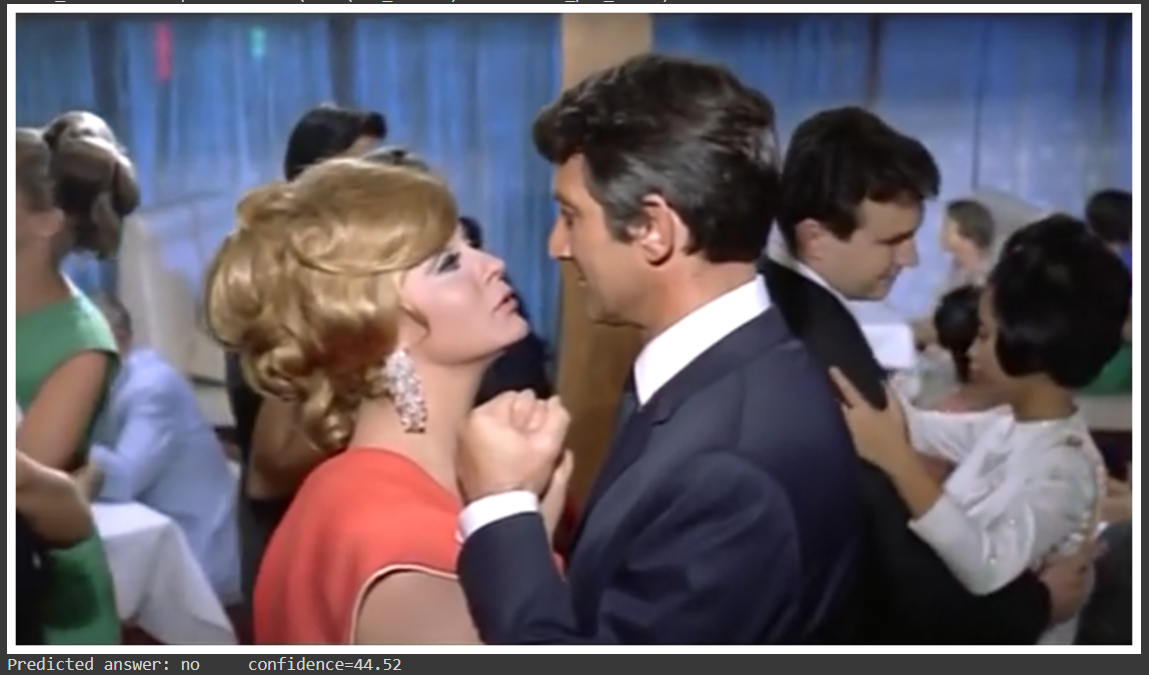} \label{fig-all-bhi}}}%
 \caption{Output of MDETR using the visual question answering used to detect and action. On the left, successful results can be observed. In this case, an image and the question ``is someone sitting?'' are given to the model. On the right, the image and the question ``is someone dancing?'' were given to the model. The example on the left shows a failure in the action detection.}%
 \label{figure-all-methods}%
\end{figure}

Figure \ref{figure-all-methods} shows a correct action detection (left) and an incorrect result (right). In this case, the frames belong to the AVA dataset and the model used is MDETR. A frame and a question about the action to detect are given to the model as input. As output, the model provides an answer, its confidence, and the location in the image where the answer was found. 

State of the art results show that the action detection task is far from solved. Up to our best knowledge, the best performing model achieved 38.8 mAP \cite{wei2022masked}. Since in our experiments we are using the standard MDETR model, not new training was required. In our case, the overall performance of the model is orders of magnitude below the best reported results. This was an expected outcome. 

Table \ref{tabe-results} shows quantitative results where MDETR performed the best and where it did not detect actions. Depending on the point of view the results can be interpreted as negative or positive. The negative aspect is that the model cannot reach results as the models designed to achieve the task specifically. The positive aspect is that MDETR is detecting actions without any additional training. This is a remarkable fact considering that the original design was targeting other tasks.

\begin{table}
\centering
\caption{Table captions should be placed above the tables.}
\label{tabe-results}
\begin{tabular}{|cccc|}
\hline
\multicolumn{4}{|c|}{\textbf{Pascal Boxes Categories Results}}                                            \\ \hline
\multicolumn{2}{|c|}{\textbf{Best Performance}}                & \multicolumn{2}{c|}{\textbf{Worst Performance}}            \\ \hline
\multicolumn{1}{|l|}{\textbf{Category}} & \multicolumn{1}{c|}{\textbf{AP@0.5IOU}} & \multicolumn{1}{c|}{\textbf{Category}} & \textbf{AP@0.5IOU} \\ \hline
\multicolumn{1}{|l|}{sleep}       & \multicolumn{1}{c|}{0.0019}     & \multicolumn{1}{l|}{answer phone}          & 0.0       \\ 
\multicolumn{1}{|l|}{sit}        & \multicolumn{1}{c|}{0.0016}     & \multicolumn{1}{l|}{kiss (a person)}        & 0.0       \\ 
\multicolumn{1}{|l|}{stand}       & \multicolumn{1}{c|}{0.0011}     & \multicolumn{1}{l|}{throw}             & 0.0       \\ 
\multicolumn{1}{|l|}{hand shake}    & \multicolumn{1}{c|}{0.0005}     & \multicolumn{1}{l|}{touch (an object)}       & 0.0       \\ 
\multicolumn{1}{|l|}{dance}       & \multicolumn{1}{c|}{0.0003}     & \multicolumn{1}{l|}{write}             & 0.0       \\ \hline
\end{tabular}
\end{table}

\section{Conclusions and Future Work}
\label{section-conlusions}

In this work we showed the use of an end-to-end text and image understanding transformer model in a task that it was not designed for. We obtained quantitative results using a challenging action recognition dataset and we tested the limits of the architecture. The remarkable characteristic that makes MDETR unique is that the model can infer over classes that it did not see before. For instance, it can detect a pink elephant (not present in the annotations). We wanted to push this aspect to this limit in the case of action detection. Although the model achieves poor quantitative results, it is possible to detect actions. This is an outstanding achievement considering the scenario of experiments.

As future work, we plan to train MDETR in the action detection task. We understand that there is potential in this line of research. Since the AVA dataset provides a high number of labels, the task seems feasible. We believe that there is room for generalization in the use of multi-modal transformers models.

\newpage

\bibliographystyle{splncs04}
\bibliography{references}
\end{document}